\documentclass[
]{ceurart}
\usepackage{listings}
\begin{document}

\copyrightyear{2022}
\copyrightclause{Copyright for this paper by its authors.
  Use permitted under Creative Commons License Attribution 4.0
  International (CC BY 4.0).}

\conference{ASAIL 2026: 8th Workshop on Automated Semantic Analysis of Information in Legal Text, June 08--12, 2026, Singapore}

\title{Chunking German Legal Code}

\author[1]{Max Prior}[%
orcid=0009-0005-7066-996X,
email=max.prior@tum.de,
]
\address[1]{Technical University of Munich, Boltzmannstraße 3, 85748 Garching near Munich}

\author[1]{Natalia Milanova}[%
orcid=0009-0003-5644-6466,
email=natalia.milanova@tum.de,
]

\author[1]{Andreas Schultz}[%
orcid=0009-0002-2893-748X,
email=andreas.schultz@tum.de,
]


\begin{abstract}
This paper investigates chunking strategies for retrieval-augmented generation on German statutory law, using the German Civil Code as a structured benchmark corpus. We implement and compare a range of segmentation approaches, including structural units (sections, subsections, sentences, propositions), fixed-size windows, contextual chunking, semantic clustering, Lumber-style chunking, and RAPTOR-based hierarchical retrieval. All methods are evaluated on a legal question-answering dataset with section-level gold labels, measuring recall, query latency, index build time, and storage requirements. Results show that chunking strategies aligned with the inherent legal structure — particularly section and subsection - based retrieval—achieve the highest recall, while more complex approaches that override this structure perform worse. These simpler methods also offer favorable computational efficiency compared to LLM-intensive techniques such as contextual chunking, RAPTOR, and Lumber. The findings highlight a key trade-off between semantic enrichment and operational cost, and demonstrate that preserving domain-specific structure is critical for effective legal information retrieval.
\end{abstract}

\begin{keywords}
  Large Language Models \sep
  Chunking \sep
  Ranking \sep
  Retrieval \sep
  Legal QA
\end{keywords}

\maketitle

\section{Introduction}
Large Language Models (LLMs) are increasingly utilized in legal services. According to the 2025 Thomson Reuters report on generative AI \cite{thomsonreuters2025genai_profservices}, the adoption of LLM-powered tools in legal practice nearly doubled, growing from 14\% in 2024 to 26\% in 2025. Among active users, 40\% rely on these tools daily, heavily favoring applications such as legal research (73\%).

This growing adoption motivates research into optimizing the LLM user experience, particularly the mitigation of hallucinations. Retrieval-Augmented Generation (RAG) systems address this issue by supplying models with relevant supporting documents \cite{gao2024retrievalaugmentedgenerationlargelanguage}. Redundant information clutters limited context windows and distracts the model. Furthermore, overloading the context often leads to the "lost in the middle" phenomenon \cite{liu2024lost}, where the model disproportionately weighs information at the beginning and end of a prompt while partially ignoring details from the middle. Keeping the retrieved context small and relevant mitigates this problem.

Effective chunking provides a solution to this retrieval challenge. By dividing documents into smaller, embeddable segments, these chunks act as keys to retrieve the full, relevant context for generation. To evaluate chunking on German legal code we utilize the German Civil Code (\textit{Bürgerliches Gesetzbuch}, BGB). The BGB is the central codification of German private law, in force since 1900, and covers core areas such as contract law, property law, family law, and inheritance law. Its long-standing structure, broad scope, and central role in legal practice make it a representative set for chunking strategies on codified statutory text. We partition the corpus of the BGB into section units that serve as the target documents for retrieval, and divide these sections into finer indexing chunks. An index is a searchable database of dense vector embeddings that represent the segmented legal texts. It enables the retrieval system to efficiently identify and extract the most semantically relevant text units for a given query via similarity search. We evaluate state of the art chunking methods against a German legal question-answering dataset \cite{buttner-habernal-2024-answering}.

Chunking introduces tradeoffs. Compared to relying on a model's parametric knowledge, integrating retrieval processes increases latency, negatively impacting the online user experience. In addition, the offline construction of these indexes is computationally lengthy and varies substantially across different chunking methodologies. Beyond latency and build time, the resulting indexes require persistent memory storage. Given the recent surge in disk storage prices \cite{james2025ai}, minimizing the space consumption of these retrieval indexes has become a critical operational factor. Consequently, we evaluate the retrieval correctness, latency, and storage requirements linked to each chunking method.

\subsection*{Contributions}
First, we implement and adapt state-of-the-art chunking methods tailored to the German law. Second, we conduct experiments measuring retrieval correctness, query latency, and space consumption. Our analysis reveals significant differences in retrieval performance, providing insights for balancing accuracy and efficiency in legal retrieval architectures.

\section{Background}
\label{sec:background}

\subsection{Retrieval Augmented Generation}\label{subsec:rag}
In Retrieval-Augmented Generation, the input is encoded to retrieve the top-$k$ fixed chunks from a dense vector index via maximum inner-product search, meaning the indexed and retrieved units are identical \cite{NEURIPS2020_6b493230}. Moving beyond traditional fixed-size segmentations, \cite{duarte2024lumberchunkerlongformnarrativedocument} introduces LumberChunker, which uses an LLM to dynamically identify semantic boundaries and produce variable-length chunks that better capture coherent, contextually complete ideas. While LumberChunker optimizes chunk boundaries, \cite{chen-etal-2024-dense} investigate retrieval granularity itself, demonstrating that indexing finer-grained atomic units like sentences and propositions improves retrieval accuracy and downstream QA performance by providing higher information density. To manage the resulting search space of varied or fine-grained chunks, Dense Hierarchical Retrieval (DHR) introduces a two-stage pipeline where a document-level retriever prunes irrelevant documents before a passage-level retriever reranks the evidence \cite{liu-etal-2021-dense-hierarchical}. Generalizing this hierarchical approach, RAPTOR recursively clusters and summarizes semantically related chunks bottom-up into a tree, enabling retrieval across multiple levels of abstraction – from fine-grained leaves to high-level summaries – which is especially effective for long documents and complex global queries \cite{sarthi2024raptorrecursiveabstractiveprocessing}.

\subsection{Computational and Storage Trade-offs}
While advanced segmentation methods like contextual chunking \cite{anthropic_contextual_retrieval_2024} and dynamic boundary detection \cite{duarte2024lumberchunkerlongformnarrativedocument} improve retrieval accuracy, they introduce substantial computational overhead. Processing large legal corpora with LLMs during the offline indexing phase significantly increases build time and space consumption. When processing queries, multi-stage retrieval pipelines and hierarchical tree traversals can elevate online query latency compared to flat maximum inner-product search. Evaluating a retrieval system therefore requires balancing the semantic richness of the index against its operational efficiency and storage footprint.

\subsection{Challenges in Legal Information Retrieval}\label{subsec:challanges}
Statutory laws, such as the BGB, differ from general-domain corpora in ways that pose structural and semantic challenges for standard retrieval systems.

First, statutory texts are organized along a deliberate hierarchical structure. The specific structure differs depending on the corpus. The BGB is divided into books (\textit{Bücher}), divisions (\textit{Abschnitte}), titles (\textit{Titel}), and subtitles (\textit{Untertitel}). It contains about 2,400 sections (\textit{Paragraphen}) each denoted by "§". Each section is typically composed of one or more subsections (\textit{Absätze}) and sentences (\textit{Sätze}). For simplification purposes, we do not further break down the structure of a section (e.g. into subsentences, numbers, literals). A single section typically contains one or more legal norms, structured as a set of conditions (\textit{Tatbestand}) leading to a legal consequence (\textit{Rechtsfolge}). § 433 BGB, for example, distributes the obligations arising from a contract of sale across two subsections: the seller's duties in subsection 1 and the buyer's duties in subsection 2. Subsections within a section also commonly separate the basic rule from exceptions, qualifications, or procedural variants. For example, subsection 1 of § 122 BGB establishes that a person who successfully avoids a declaration of intent must compensate the other party for reliance damages, while subsection 2 excludes this liability where the injured party knew or should have known of the ground for avoidance.

Second, statutory provisions are densely cross-referenced, and the references take several forms. Explicit references name the target provision directly, as in § 437 BGB, which lists the buyer's remedies in case of defects by referring to – \textit{inter alia } – §§ 439, 440, 441, 323, and 326 BGB. Implicit references invoke a target provision through legal terminology rather than by citation. For instance, a reference to "consumer" presupposes the legal definition in § 13 BGB without naming it. References may further be internal, pointing to other provisions within the same corpus (here the BGB), or external, pointing to provisions in another corpus. Intra-section references, in which one subsection refers to another subsection of the same section, are also common.

Third, statutory provisions use feature-dense and highly specific language. Single words can carry decisive legal weight. For example, the difference between \textit{unverzüglich} ("without undue delay") and \textit{sofort} ("immediately") changes the legal consequence of a provision such as § 121 BGB regarding the period of avoidance of certain – otherwise legally binding – declaration of intents.

Traditional fixed-size chunking often severs these precise legal boundaries and leads to a critical loss of context. Consequently, specialized segmentation strategies that respect natural statutory boundaries, such as subsections and individual legal propositions, are necessary to preserve as much of the legal meaning as possible for accurate downstream question answering.








\section{Chunking Strategies}

This section describes the chunking strategies evaluated in our experiments. We first introduce structure-preserving baselines that split the BGB into sections, subsections, sentences, or propositions. We then compare these baselines with methods that modify or aggregate these units, including fixed-size windows, Lumber-style, contextual chunking, semantic clustering, and RAPTOR-based hierarchical retrieval. Across all strategies, the indexed units may differ, but evaluation is always performed at the parent-section level: retrieved chunks, clusters, or summary nodes are mapped back to the BGB sections from which they originate.

\paragraph{Subsection, sentence and fixed size splitting} 
In the following, the decomposition of § 535 BGB (contents and primary duties of the lease agreement) on fixed-size, subsection, and sentence level is illustrated. (1) and (2) indicate the start of a subsection; S marks sentence chunks. Unlike the conventional German legal citation style, we number sentences continuously across the section to reflect the chunking system's internal representation.

\begin{quote}\footnotesize
\textbf{(1)} \textbf{S1} A lease agreement imposes on the lessor a duty to grant the lessee use of the leased property for the lease period. \textbf{S2} The lessor is to make available the leased property to the lessee in a condition suitable for use as contractually agreed and maintain it in this condition for the lease period. \textbf{S3 }The lessor is to bear all costs to which the leased property is subject.

\textbf{(2)} \textbf{S4} The lessee is obliged to pay the lessor the agreed rent.
\end{quote}
Subsection chunking returns exactly the two blocks. Sentence chunking returns the four tagged units \textbf{S1}--\textbf{S4}. Fixed-size chunking ignores these legal boundaries. In a fixed-size setting, the same provision is captured schematically as overlapping fixed length windows \textbf{F1} - \textbf{F2}:



\begin{quote}\footnotesize
\textbf{F1} \textbf{(1)} \textbf{S1} A lease agreement imposes on the lessor a duty to grant the lessee use of the leased property for the lease period. \textbf{S2} The lessor is to make available the leased property to the lessee in a condition suitable for use as contractually agreed and maintain it in this

\textbf{F2} in a condition suitable for use as contractually agreed and maintain it in this condition for the lease period. \textbf{S3} The lessor is to bear all costs to which the leased property is subject. \textbf{(2)} \textbf{S4} The lessee is obliged to pay the lessor the agreed rent.
\end{quote}
This splitting method guarantees uniform chunk sizes but frequently severs legal contexts mid-sentence, forcing the retrieval system to rely on the sliding overlap to capture complete statements across adjacent windows.

\paragraph{Proposition chunking}
Proposition chunking \cite{chen-etal-2024-dense} decomposes – to the extent possible – a subsection or sentence into smaller, self-contained statements. The example below is taken from \S~433 (1) S1 BGB (contractual duties typical for a purchase agreement). If no further decomposition is possible, the sentence or subsection remains unchanged.

\begin{quote}\footnotesize
\textbf{Original sentence:} \\
By a purchase agreement, the seller of a thing is obliged to deliver the thing to the buyer and to procure ownership of the thing for the buyer.

\textbf{Proposition 1:} \\
By the purchase agreement, the seller of a thing is obliged to deliver the thing to the buyer.

\textbf{Proposition 2:} \\
By the purchase agreement, the seller of a thing is obliged to procure ownership of the thing for the buyer.
\end{quote}

The sentence is split into two propositions, each expressing a single legal obligation. Generally speaking, each proposition retains the conditions of the original norm and isolates one of its legal consequences. If a sentence expresses only a single self-contained statement, it remains unchanged and the resulting proposition coincides with the sentence itself. All experiments are conducted with three indexing granularities: subsection, sentence, and proposition. Unless stated otherwise, each method is evaluated separately for all three. In the following, we refer to these as \emph{base units}.

\paragraph{Lumber-style chunking}
Lumber-style chunking \cite{duarte2024lumberchunkerlongformnarrativedocument} converts the text into a sequential stream of base units (e.g., propositions). Starting at the current position, units are accumulated up to fixed upper bound of tokens and passed to an LLM, which predicts the first unit where the \emph{next} chunk should begin. All preceding units form one chunk; if no earlier split is predicted, the full group is kept. The result is a dynamic multi-unit chunk, while retrieval expands back to their parent sections. As these chunks can span across several retrieval units, a single retrieved unit may map to multiple distinct parent sections.

An example taken from \S\S~111--113 BGB yields two chunks, with a boundary that cuts across \S~112:







\begin{quote}\footnotesize
\textbf{Lumberchunk L1}

\textbf{§ 111 BGB (Unilateral legal transactions)}

\textbf{S1 }A unilateral legal transaction that a minor undertakes without the necessary consent of the legal representative is ineffective.
\textbf{S2} If the minor undertakes such a legal transaction with regard to another person with this consent, the legal transaction is ineffective if the minor does not present the consent in writing and the other person rejects the legal transaction for this reason without undue delay.
\textbf{S3} Rejection is not possible if the representative had given the other person notice of the consent.

\textbf{§ 112 (1) BGB (Independent operation of a trade or business)}

\textbf{S4} If the legal representative, with the ratification of the family court, authorises the minor to operate a trade or business independently, the minor has unlimited capacity to contract for such transactions as the business operations entail.

\textbf{Lumberchunk L2}

\textbf{§ 112 (1) BGB}

\textbf{S5} Legal transactions are exempt for which the representative needs to obtain the ratification of the family court.

\textbf{§ 112 (2) BGB}

\textbf{S6 }The authorisation may be revoked by the legal representative only with the ratification of the family court.

\textbf{§ 113 (1) BGB (Service or employment relationship)}

\textbf{S7} If the legal representative authorises the minor to enter service or employment, the minor has unlimited capacity to enter into transactions that relate to entering or leaving service or employment of the permitted nature or performing the duties arising from such a relationship.

\textbf{S8 }Contracts are exempt for which the legal representative needs to obtain the ratification of the family court.

\textbf{§ 113 (2) BGB}

\textbf{S9} The authorisation may be revoked or restricted by the legal representative.
\end{quote}

The sentences \textbf{S1}--\textbf{S9} serve only as the segmentation basis; the indexed units are the two chunks \textbf{L1} and \textbf{L2}, which span \S\S~111--112 and \S\S~112--113, respectively. The boundary therefore separates the rule of § 112(1) — that the minor has unlimited capacity for business transactions — from its immediate exception, which restricts that capacity for transactions requiring family-court approval. From the lawmaker's perspective, these two sentences are intended to be read together.

\paragraph{Contextual chunking}
Contextual chunking augments base units with a short LLM-generated context before embedding, inspired by the contextual-embeddings idea of contextual retrieval \cite{anthropic_contextual_retrieval_2024}. In our implementation, the base unit and its surrounding additional context are processed by an LLM to generate a concise contextual prefix. This prefix is then prepended to the base unit, and the resulting concatenated string is embedded to serve as the indexing unit. Only the embedding text is modified; retrieval is still aggregated back to the parent section.

\begin{quote}\footnotesize
\textbf{Base unit (S4 from § 535 BGB):} \\
The lessee is obliged to pay the lessor the agreed rent.

\vspace{0.5em}
\textbf{Additional context (hierarchy + full section):}
Book 2: Law of Obligations (\textit{Schuldrecht})
Division 8: Particular Types of Obligations (\textit{Einzelne Schuldverhältniss}e)
Title 5: Lease (\textit{Mietvertrag, Pachtvertrag)}
Subtitle 1: Lease (\textit{Mietvertrag})
Section 535: Contents and primary duties of the lease agreement 
(\textit{Inhalt und Hauptpflichten des Mietvertrags})

Section 535
Contents and primary duties of the lease agreement

(1) A lease agreement imposes on the lessor a duty to grant the lessee use of the leased property for the lease period. The lessor is to make available the leased property to the lessee in a condition suitable for use as  contractually agreed and maintain it in this condition for the lease period. The lessor is to bear all costs to which the leased property is subject.

(2) The lessee is obliged to pay the lessor the agreed rent.

\vspace{0.5em}
\textbf{Contextual prefix:} \\
Lease law: states the lessee's primary payment obligation under \S~535 BGB.

\textbf{Embedded contextualized unit:} \\
Additional context: Lease law: states the lessee's primary payment obligation under \S~535 BGB. \\
Sentence: The lessee is obliged to pay the lessor the agreed rent.
\end{quote}

\paragraph{Semantic clustering}
In this method, we group semantically similar base units within the BGB\footnote{For each experiment, the base unit is fixed to one granularity: subsection, sentence, or proposition. Granularities are not mixed; if a section cannot be split further, the resulting unit remains identical to the original section, subsection, or sentence.}. We apply the KMeans \cite{hartigan1979kmeans} algorithm to form the clusters.

\begin{quote}\footnotesize
  \textbf{\S~107 (Consent of legal representative):} \\
 For a declaration of intent as a result of which minors do not receive merely a legal benefit, the minors require consent by their legal representative.

  \textbf{\S~108 (1) (Entry into a contract without consent):} \\
  If the minor enters into a contract without the necessary consent of the legal representative, the effectiveness of the contract is subject to approval by the legal representative.

  \textbf{\S~111 sentence~1 (Unilateral legal transactions):} \\
  A unilateral legal transaction that a minor undertakes without the necessary consent of the legal representative is ineffective
\end{quote}

While this example uses sentences as the base unit, we conduct identical experiments using subsections and propositions. In this sentence-based variant, all BGB sentences are initially embedded and then clustered based on cosine similarity. The embeddings within each cluster are pooled into a single vector, which becomes the indexing unit. In this example, the indexed unit is the pooled vector of a cross-section semantic cluster containing sentences from \S\S~107, 108, and 111. A retrieved cluster embedding maps back to all of its parent sections.
  
\paragraph{RAPTOR}
Drawing on the core concept of RAPTOR-style hierarchical retrieval \cite{sarthi2024raptorrecursiveabstractiveprocessing} the base units are embedded as leaf nodes, clustered bottom-up, summarized by an LLM, and recursively re-embedded to form a hierarchy of summary nodes. At query time, retrieval is routed through these summary nodes and then descends to relevant leaves, which are aggregated back to their parent sections.

The example comes from \S\S~339--341 BGB regarding contractual penalties:

\begin{quote}\footnotesize
  \textbf{Leaf 1 – \S~339 sentence~1 (\textbf{Payability of penalty for breach of contract}):} \\
  Where the obligor promises the obligee, in the event of their failing to perform their obligation or failing to do so properly, payment of an amount of money as a penalty, the penalty is payable upon the obligor being in default.

  \textbf{Leaf 2 – \S~340(1) sentence~1 (Promise to pay a penalty for non-performance):} \\
  If the obligor has promised the penalty in the event of their failing to perform their obligation, then the obligee may demand the penalty that is payable in lieu of fulfilment.

  \textbf{Leaf 3 – \S~341(1) (Promise of a penalty for improper performance):} \\
  If the obligor has promised the penalty in the event of their failing to perform their obligation properly, including performance at the specified time, the obligee may demand the payable penalty in addition to performance.

  \textbf{Root summary node:} \\
  The provisions of \S\S~339--341 BGB regulate contractual penalties, including forfeiture conditions and whether penalties replace or accompany performance.
\end{quote}

The leaves capture related consequences of contractual penalties and are grouped under a shared summary node. The indexed structure is the embedded summary, while retrieval returns the corresponding parent sections.

\section{Data and Experiments}
We evaluate on 525 validation questions with section-level gold labels from \texttt{DomainLLM/gerlayqa-bgb-paraphrased} \cite{buttner-habernal-2024-answering} against the full BGB corpus, which is normalized into 2,455 hierarchical sections. All methods share this normalized section and sentence representation, with propositions generated once via \texttt{minimax-m2.5} and similarity search embeddings computed using \texttt{gemini-embedding-001}.  The \texttt{gerlayqa-bgb-paraphrased} dataset challenges the retrieval system to map everyday problem descriptions to statutory provisions. An evaluation sample from \cite{buttner-habernal-2024-answering} takes the following form:

\begin{quote}\footnotesize
\textbf{Example Query (Paraphrased Layperson Question):} \\
An acquaintance terminated her apartment lease in Frankfurt am Main effective December 31, 2003, and moved to Berlin. After a change of landlord, there were difficulties with the refund of the security deposit. By when can she demand the return of the deposit and, if necessary, sue for it?

\textbf{Gold Section Labels (Relevant BGB Sections):} \\
\S~ 548 (Limitation of compensation claims and of the right of removal)\\
\S~ 566a (Rent security deposit)
\end{quote}

\begin{figure}[!ht]
\centering
 \includegraphics[width=1\linewidth]{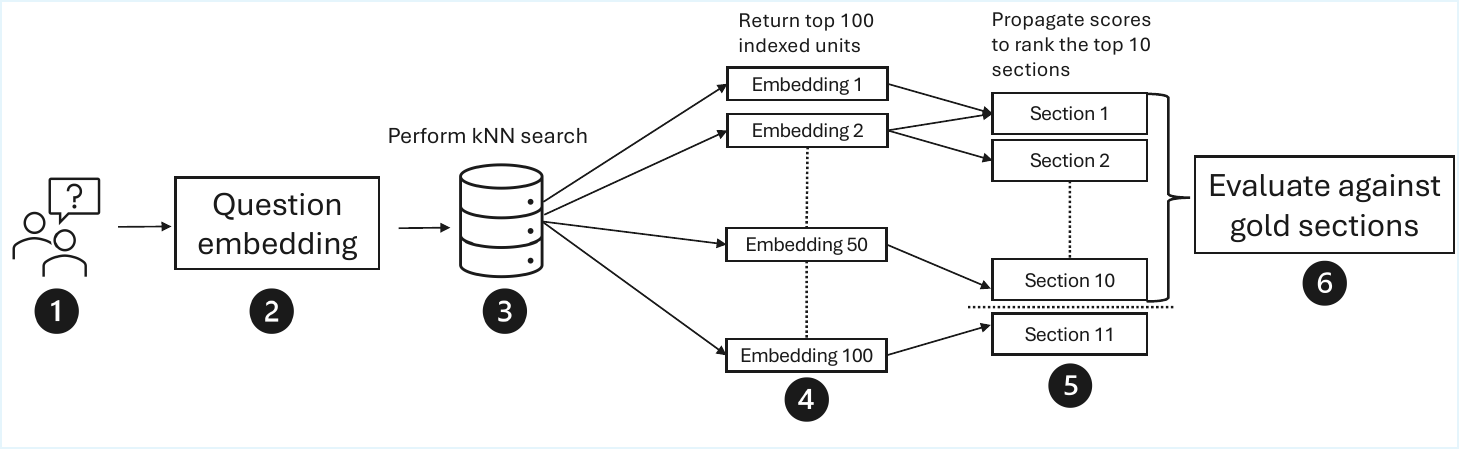}
 \caption{Query Processing}
\label{fig:query_processing}
\end{figure}

Figure~\ref{fig:query_processing} outlines the query processing pipeline applied across all methods. A layperson query is embedded first (Steps 1 and 2), and a kNN search is done over the respective index (Step 3). Regardless of the underlying indexing unit, the system first retrieves a candidate set of the top 100 embedded indexed units, such as chunks, leaves, or clusters (Step 4). Unlike base unit and contextual embeddings, which map to a single parent section, other indexed units can map to multiple. Fixed-size and Lumber chunks bridge adjacent sections, semantic clusters pool sentences from adjacent and non-adjacent sections, and RAPTOR summary nodes map to all parent sections of their constituent leaves. These embedding similarity scores are then propagated to their parent sections for ranking (Step 5). The top 10 distinct sections with the highest aggregated scores form the final result, which is then evaluated against the gold standard sections (Step 6).

We measure the four mentioned metrices for evaluation: recall, which serves as the primary metric by comparing the top 10 retrieved distinct sections against the gold standard labels; query latency, which assesses the online search and aggregation cost by measuring the time required to retrieve 100 indexing units and rank the parent sections while excluding the initial query embedding time; offline build time, which accounts for the total time required to construct the retrieval index; and space consumption, which quantifies the persistent storage footprint of the retrieval-ready index and its metadata, excluding the raw text payloads.

\paragraph{Recall}

\begin{figure}[!ht]
\centering
 \includegraphics[width=1\linewidth]{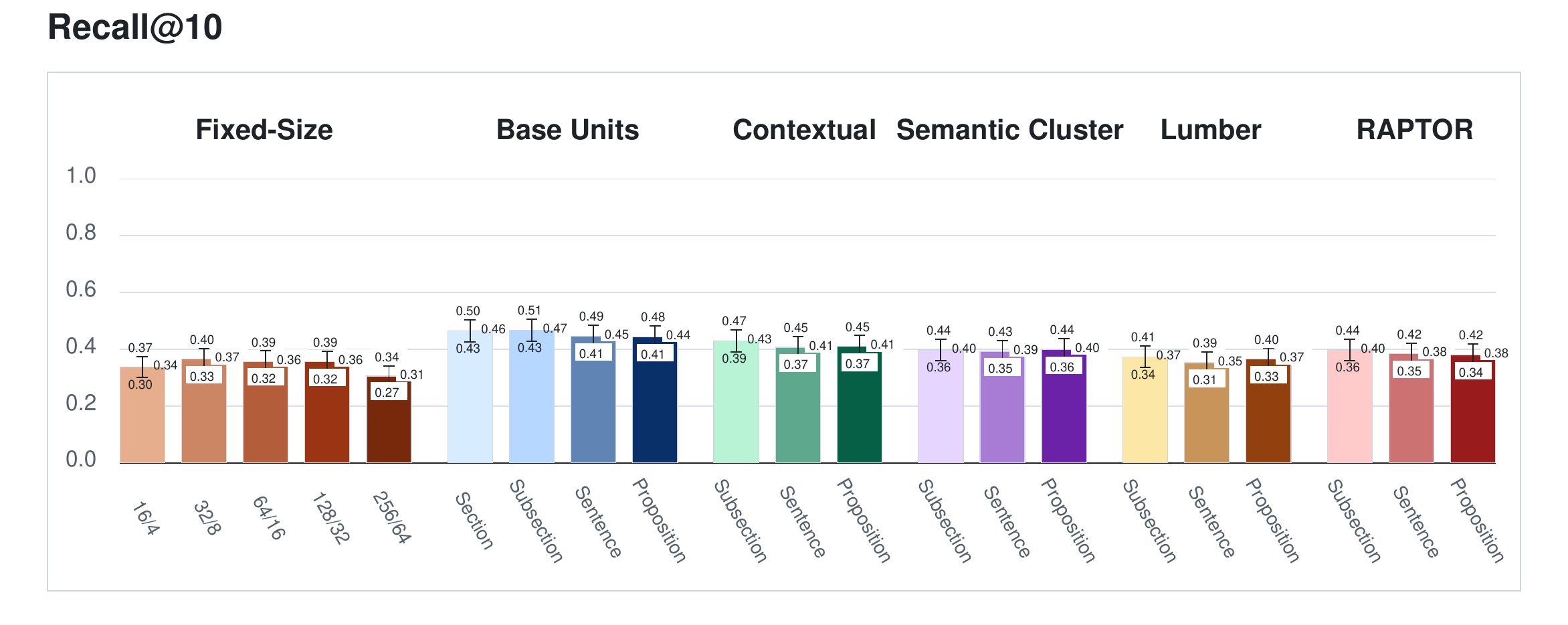}
 \caption{Recall@10}
\label{fig:q525_columns_recall}
\end{figure}

Figure~\ref{fig:q525_columns_recall} reports the mean section-level Recall@10 for the full benchmark. It compares 21 retrieval strategies on 525 validation questions. On average each question has approximately 1.7 gold sections for the answers. The bars show the mean across questions with 95\% confidence intervals. The results indicate that the strongest recall is achieved by methods that preserve the native legal structure: subsection retrieval performs best (0.47), followed very closely by section retrieval (0.46), while sentence and proposition retrieval form the next strongest group. Contextual retrieval remains competitive but clearly below these structural baselines, and the semantic-cluster and RAPTOR variants occupy a middle tier. Lumber and fixed-size chunking perform worst overall, with the coarsest fixed-size setting, Fixed 256 / 64, producing the lowest recall.

We use a repeated-measures testing setup because all methods are evaluated on the same 525 questions. First, we apply Friedman's non-parametric test \cite{Friedman01121937} as an omnibus test: it only checks whether there is any overall evidence that not all methods have the same Recall@10, without identifying which methods differ. This is appropriate for matched per-question method scores and does not assume normally distributed Recall@10 values. Since the Friedman test rejects equality across methods ($p<0.0001$), we then test the specific comparisons of each method against section retrieval using paired permutation tests. Because section retrieval is compared with many other methods, we apply Holm correction \cite{holm1979simple} to reduce the risk of declaring a difference significant by chance. The corrected pairwise results show no significant difference between section retrieval and the other structural base units, but section retrieval is significantly better than all other strategies.

Overall, the results suggests that, under a strict top-10 section evaluation, preserving statutory structure at section or subsection level is more beneficial for retrieval quality than using more complex global or fixed-window chunk organizations. Ablation experiments with a top-25 section evaluation confirm this result.

\paragraph{Query performance and build time}

\begin{figure}[!ht]
\centering
\includegraphics[width=1\linewidth]{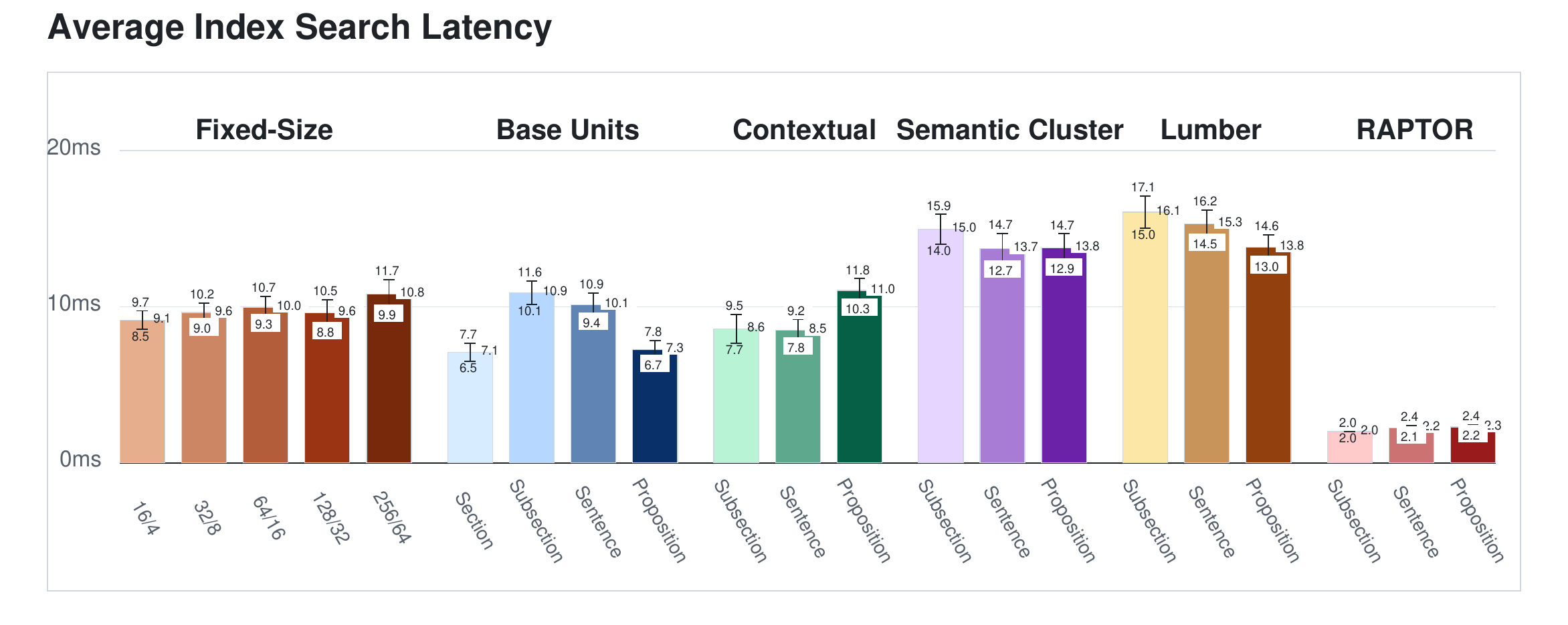}
\caption{Average Index Search Latency}
\label{fig:q525_columns_latency}
\end{figure}

Figure~\ref{fig:q525_columns_latency} reports mean retrieval-only latency, averaged over 5 runs, isolating the cost of kNN vector search and section aggregation while excluding query embedding and text materialization. Strategy-level differences are primarily architectural. RAPTOR is the fastest (2.00--2.30,ms) because it prunes the search space hierarchically, avoiding exhaustive searches. Flat baseline, contextual, and fixed-size strategies are slower, as they query a single vector space before aggregating hits. Semantic clustering and Lumber are the slowest (13.72--16.06,ms) due to extensive post-retrieval consolidation across multiple covered sections. Thus, early pruning capabilities influence latency more than raw index size. Within strategies, the chosen base unit dictates evidence concentration. In flat baselines, latency scales as \texttt{section} $\approx$ \texttt{proposition} $<$ \texttt{sentence} $<$ \texttt{subsection}, since \texttt{proposition} hits map to fewer distinct parent sections, minimizing aggregation overhead. Across other strategies, unit-level variations reflect similar secondary effects of chunk selectivity. Ultimately, the unit's ambiguity during final aggregation is more decisive than the total number of indexed vectors. To evaluate latency significance on the 525 questions, we use Holm-corrected paired permutation tests with paired bootstrap 95\% confidence intervals to compare each method directly against section retrieval. Results indicate RAPTOR variants are significantly faster, most other strategies are significantly slower, and proposition retrieval is statistically indistinguishable from section retrieval.

\begin{figure}[!ht]
\centering
\includegraphics[width=1\linewidth]{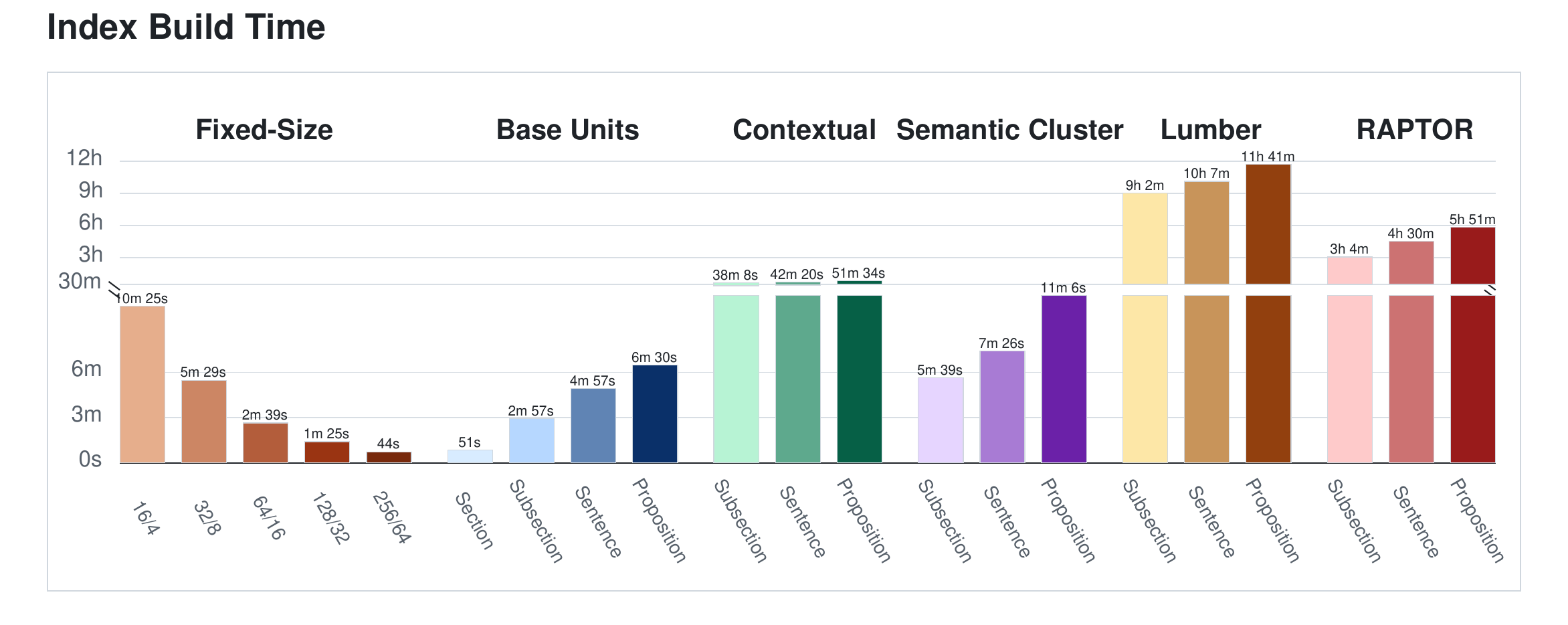}
\caption{Offline build time}
\label{fig:q525_columns_build_time}
\end{figure}

Figure~\ref{fig:q525_columns_build_time} reports the offline index build time. The cheapest strategies are deterministic flat indexes: Fixed 256 / 64 takes 44\,s and Section retrieval 51\,s because they create few chunks, while smaller fixed windows become slower as overlap creates many more vectors, rising to 10\,m 25\,s for Fixed 16 / 4. The base-unit methods remain in the minute range but grow with unit count and granularity, from subsection at 2\,m 57\,s to Sentence at 4\,m 57\,s and Proposition at 6\,m 30\,s. Semantic-cluster methods add clustering and pooled cluster vectors, so they are higher but still moderate at 5\,m 39\,s to 11\,m 6\,s, again increasing from subsection to Proposition. Contextual subsection takes 38\,m 8\,s, contextual Sentence 42\,m 20\,s, and contextual Proposition 51\,m 34\,s, because each unit first requires LLM-generated context before embedding and the finer units produce more chunks. RAPTOR is much more expensive at 3\,h 4\,m to 5\,h 51\,m because it recursively builds and embeds a summary tree requiring a substantial amount of LLM calls, and Lumber is the highest-cost family at 9\,h 2\,m to 11\,h 41\,m because it relies on long LLM-based corpus construction. Overall, build time is driven by how much generative preprocessing each index requires.

\paragraph{Disk space}

\begin{figure}[!ht]
\centering
\includegraphics[width=1\linewidth]{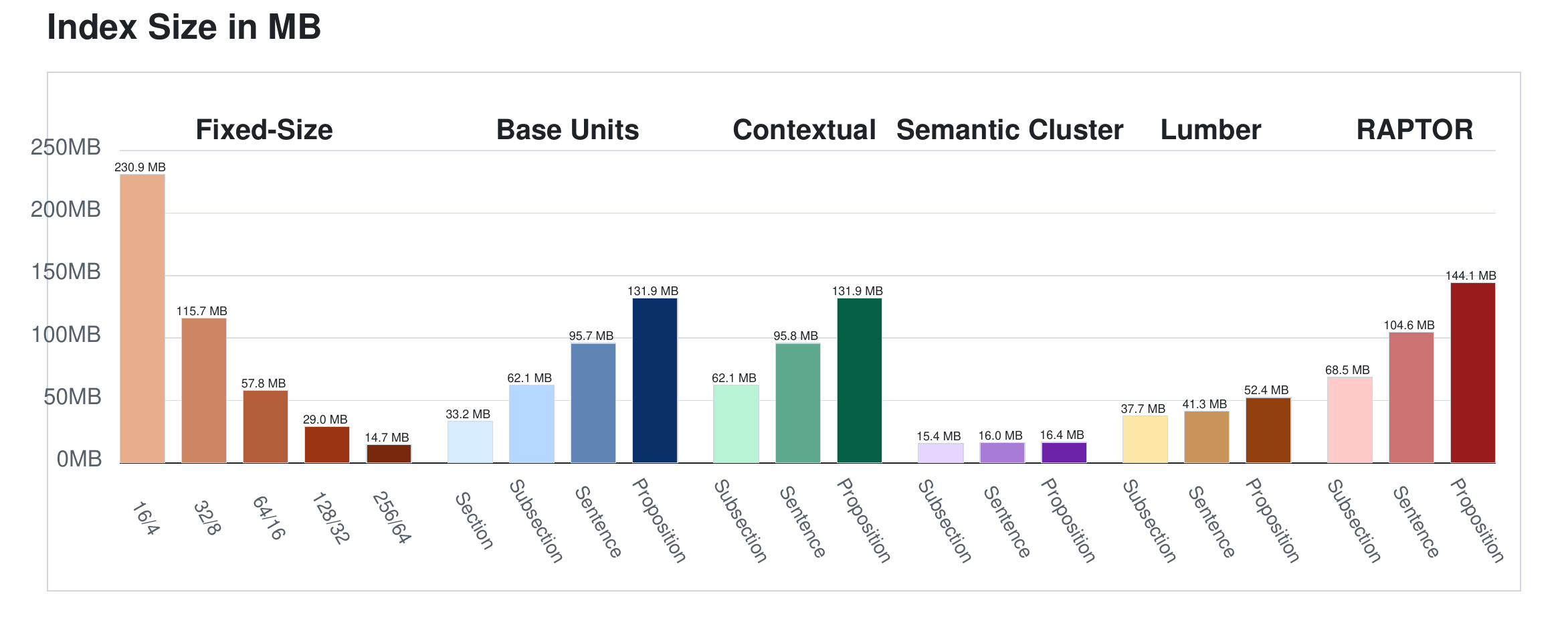}
\caption{Total Persisted Size for the plotted methods.}
\label{fig:q525_columns_space}
\end{figure}

Figure~\ref{fig:q525_columns_space} reports the corpus-level, retrieval-ready index size in MB, excluding raw text payloads. Index size is primarily determined by the final stored vector count and routing metadata rather than the strategy family alone. The most compact methods are Fixed 256 / 64 (14.7MB) and semantic clustering (15.4--16.4MB), as both aggressively reduce stored vectors. Lumber occupies the middle tier (37.7--52.4MB). Conversely, flat baseline and contextual strategies require more storage by indexing every leaf unit directly, with sizes increasing monotonically from \texttt{section} (33.2MB) to \texttt{subsection} (62.1MB), \texttt{sentence} (95.7,MB), and \texttt{proposition} (131.9,MB) due to the proliferation of finer units. Since they embed the same number of indexing units, their space requirements are identical. RAPTOR exhibits similar growth but at a higher baseline (68.5MB to 144.1MB) because it stores both leaf and summary routing nodes. Highly overlapping fixed windows, such as Fixed 16 / 4, consume the most space (230.9MB). Semantic clustering remains the notable exception to this unit-size effect, maintaining a stable footprint across all base units because its builder consistently compresses them into roughly one thousand final cluster vectors.

\section{Discussion and Conclusion}

As shown above, the strongest Recall@10 is achieved by methods that preserve the statutory structure of the BGB: subsection retrieval and section retrieval, followed by sentence and proposition retrieval. Methods that override this structure – fixed-size windows, semantic clustering, and Lumber-style boundary prediction – perform measurably worse. The deliberate structure of a legal corpus such as the BGB, as discussed in Section \ref{subsec:challanges}, may offer explanations for this finding. A legal corpus is not structured by arbitrary text spans but by boundaries drawn deliberately by the historic lawmaker to delimit regulatory matters, grouping conditions, consequences, and qualifications that are intended to be read together. Retrieval at this granularity therefore returns units that are aligned both with the legislator's segmentation and with the section-level gold labels.

Semantic clustering and Lumber-style chunking, by contrast, pool multiple provisions into a single indexed unit on the basis of thematic similarity or LLM-predicted coherence. The resulting chunk is represented by a single embedding that averages over all pooled provisions. When the answer to a specific legal question lies in one particular provision rather than across the thematic group, this averaging blurs the embedding away from the relevant provision and toward the shared topic, which may explain the weaker recall of these methods. RAPTOR occupies a middle tier. Its hierarchical summaries provide some routing benefit but, like semantic clustering, ultimately group provisions across the legislator's structural boundaries. Provisions grouped under a shared summary may share a topic but differ in legally decisive details – for instance, when one provision sets a rule and another defines an exception. Routing through such summaries does not reliably lead to the specific provision a query targets. Fixed-size chunking ignores boundaries altogether and routinely separates conditions from consequences. The sliding overlap mitigates but does not eliminate this effect.

A factor not addressed by any of the chunking strategies evaluated here is the resolution of cross-references. As discussed in Section \ref{subsec:challanges}, the BGB – like virtually any legal corpus – makes extensive use of explicit, implicit, and intra-section references; the relevant legal context for a query may therefore be distributed across multiple sections. Strategies that automatically follow such references and incorporate the referenced text into the indexed unit – for example by expanding a section to include the legal definitions it relies on – may improve recall further. We leave the implementation and empirical evaluation of such reference-aware chunking strategies to future work.





\section{Limitations}
This study relies on a single layperson dataset \cite{buttner-habernal-2024-answering} limited to the BGB, meaning the results may not generalize to expert queries or other legal frameworks. Furthermore, we evaluate only section-level retrieval rather than downstream generative performance, and our current implementation ignores cross-reference resolution while depending on specific embedding models that could influence the observed trade-offs. A failure analysis investigating the specific patterns and bottlenecks of missed documents is planned for future work.

\begin{acknowledgments}
This work was funded by the Digitalisierungsinitiative des Bundes für die Justiz under the "Generatives Sprachmodell der Justiz (GSJ)" project, a collaboration between the justice ministries of North Rhine-Westphalia and Bavaria, TU Munich, and the University of Cologne
\end{acknowledgments}

\section*{Declaration on Generative AI}
During the preparation of this work, the authors used Gemini to polish the text, and Codex in order to generate code. After using these tools, the author reviewed and edited the content as needed and take full responsibility for the publication's content.

\bibliography{sample-ceur}

@inproceedings{NEURIPS2020_6b493230,
 author = {Lewis, Patrick and Perez, Ethan and Piktus, Aleksandra and Petroni, Fabio and Karpukhin, Vladimir and Goyal, Naman and K\"{u}ttler, Heinrich and Lewis, Mike and Yih, Wen-tau and Rockt\"{a}schel, Tim and Riedel, Sebastian and Kiela, Douwe},
 booktitle = {Advances in Neural Information Processing Systems},
 pages = {9459--9474},
 publisher = {Curran Associates, Inc.},
 title = {Retrieval-Augmented Generation for Knowledge-Intensive NLP Tasks},
 url = {https://proceedings.neurips.cc/paper_files/paper/2020/file/6b493230205f780e1bc26945df7481e5-Paper.pdf},
 volume = {33},
 year = {2020}
}

@inproceedings{liu-etal-2021-dense-hierarchical,
    title = "Dense Hierarchical Retrieval for Open-domain Question Answering",
    author = "Liu, Ye  and
      Hashimoto, Kazuma  and
      Zhou, Yingbo  and
      Yavuz, Semih  and
      Xiong, Caiming  and
      Yu, Philip",
    booktitle = "Findings of the Association for Computational Linguistics: EMNLP 2021",
    month = nov,
    year = "2021",
    publisher = "Association for Computational Linguistics",
    doi = "10.18653/v1/2021.findings-emnlp.19",
    pages = "188--200"
}

@misc{duarte2024lumberchunkerlongformnarrativedocument,
      title={LumberChunker: Long-Form Narrative Document Segmentation}, 
      author={André V. Duarte and João Marques and Miguel Graça and Miguel Freire and Lei Li and Arlindo L. Oliveira},
      year={2024},
      archivePrefix={arXiv},
      primaryClass={cs.CL},
      url={https://arxiv.org/abs/2406.17526}, 
}

@misc{sarthi2024raptorrecursiveabstractiveprocessing,
      title={RAPTOR: Recursive Abstractive Processing for Tree-Organized Retrieval}, 
      author={Parth Sarthi and Salman Abdullah and Aditi Tuli and Shubh Khanna and Anna Goldie and Christopher D. Manning},
      year={2024},
      archivePrefix={arXiv},
      primaryClass={cs.CL},
      url={https://arxiv.org/abs/2401.18059}, 
}

@inproceedings{chen-etal-2024-dense,
    title = "Dense {X} Retrieval: What Retrieval Granularity Should We Use?",
    author = "Chen, Tong  and
      Wang, Hongwei  and
      Chen, Sihao  and
      Yu, Wenhao  and
      Ma, Kaixin  and
      Zhao, Xinran  and
      Zhang, Hongming  and
      Yu, Dong",
    booktitle = "Proceedings of the 2024 Conference on Empirical Methods in Natural Language Processing",
    month = nov,
    year = "2024",
    address = "Miami, Florida, USA",
    publisher = "Association for Computational Linguistics",
    doi = "10.18653/v1/2024.emnlp-main.845",
    pages = "15159--15177"
}

@misc{anthropic_contextual_retrieval_2024,
  author       = {{Anthropic}},
  title        = {Contextual Retrieval},
  year         = {2024},
  howpublished = {\url{https://www.anthropic.com/engineering/contextual-retrieval}},
  note         = {Accessed: 2026-04-08}
}

@article{hartigan1979kmeans,
  author    = {Hartigan, John A. and Wong, Manchek A.},
  title     = {A K-Means Clustering Algorithm},
  journal   = {Journal of the Royal Statistical Society: Series C (Applied Statistics)},
  volume    = {28},
  number    = {1},
  pages     = {100--108},
  year      = {1979},
  publisher = {Royal Statistical Society},
  doi       = {10.2307/2346830},
}

@article{liu2024lost,
    title = "Lost in the Middle: How Language Models Use Long Contexts",
    author = "Liu, Nelson F.  and
      Lin, Kevin  and
      Hewitt, John  and
      Paranjape, Ashwin  and
      Bevilacqua, Michele  and
      Petroni, Fabio  and
      Liang, Percy",
    journal = "Transactions of the Association for Computational Linguistics",
    volume = "12",
    year = "2024",
    address = "Cambridge, MA",
    publisher = "MIT Press",
    doi = "10.1162/tacl_a_00638",
    pages = "157--173"
}

@techreport{thomsonreuters2025genai_profservices,
  title       = {Generative AI in Professional Services Report},
  institution      = {Thomson Reuters Institute},
  year        = {2025},
  url         = {https://www.thomsonreuters.com/content/dam/ewp-m/documents/thomsonreuters/en/pdf/reports/2025-generative-ai-in-professional-services-report-tr5433489-rgb.pdf},
}

@misc{gao2024retrievalaugmentedgenerationlargelanguage,
      title={Retrieval-Augmented Generation for Large Language Models: A Survey}, 
      author={Yunfan Gao and Yun Xiong and Xinyu Gao and Kangxiang Jia and Jinliu Pan and Yuxi Bi and Yi Dai and Jiawei Sun and Meng Wang and Haofen Wang},
      year={2024},
      eprint={2312.10997},
      archivePrefix={arXiv},
      primaryClass={cs.CL},
      url={https://arxiv.org/abs/2312.10997}, 
}

@misc{james2025ai,
  author = {James, Luke},
  title = {{AI} data centers are swallowing the world's memory and storage supply, setting the stage for a pricing apocalypse that could last a decade},
  year = {2025},
  howpublished = {\url{https://www.tomshardware.com/pc-components/storage/perfect-storm-of-demand-and-supply-driving-up-storage-costs}},
  note = {Accessed: 2026-04-23}
}

@inproceedings{buttner-habernal-2024-answering,
    title = "Answering legal questions from laymen in {G}erman civil law system",
    author = {B{\"u}ttner, Marius  and
      Habernal, Ivan},
    booktitle = "Proceedings of the 18th Conference of the European Chapter of the Association for Computational Linguistics (Volume 1: Long Papers)",
    month = mar,
    year = "2024",
    doi = "10.18653/v1/2024.eacl-long.122",
    pages = "2015--2027",
}

@article{Friedman01121937,
author = {Milton Friedman},
title = {The Use of Ranks to Avoid the Assumption of Normality Implicit in the Analysis of Variance},
journal = {Journal of the American Statistical Association},
volume = {32},
number = {200},
pages = {675--701},
year = {1937},
publisher = {Taylor \& Francis},
doi = {10.1080/01621459.1937.10503522}
}

@article{holm1979simple,
 ISSN = {03036898, 14679469},
 URL = {http://www.jstor.org/stable/4615733},
 author = {Sture Holm},
 journal = {Scandinavian Journal of Statistics},
 number = {2},
 pages = {65--70},
 publisher = {[Board of the Foundation of the Scandinavian Journal of Statistics, Wiley]},
 title = {A Simple Sequentially Rejective Multiple Test Procedure},
 urldate = {2026-04-26},
 volume = {6},
 year = {1979}
}

\appendix

\end{document}